\documentclass[letterpaper, 10 pt, conference]{ieeeconf}  

\IEEEoverridecommandlockouts                              

\usepackage{graphicx}
\usepackage{subcaption}
\usepackage{url}
\usepackage{booktabs}
\usepackage{amsmath}
\usepackage{algorithm}
\usepackage[noend]{algpseudocode }
\usepackage{wrapfig}
\usepackage{dramatist}
\usepackage{listings}
\usepackage{pgfplots}
\pgfplotsset{compat=1.17}
\title{\LARGE \bf
Dobby: A Conversational Service Robot Driven by GPT-4 
}

\author{Carson Stark, Bohkyung Chun, Casey Charleston, Varsha Ravi,\\Luis Pabon, Surya Sunkari, Tarun Mohan, Peter Stone, and Justin Hart\thanks{Department of Computer Science, The University of Texas at Austin, Austin, Texas, USA 78712
{\tt\small {\{pstone, hart\}@cs.utexas.edu
\{carsonstark, boh, caseycharleston, vravi, luisalepabon, suryasunkari, tarun.mohan\}@utexas.edu
}} Peter Stone is also with Sony AI.
}}

\begin{document}

\maketitle
\thispagestyle{empty}
\pagestyle{empty}

\begin{abstract}

This work introduces a robotics platform which embeds a conversational AI agent in an embodied system for natural language understanding and intelligent decision-making for service tasks; integrating task planning and human-like conversation. The agent is derived from a large language model, which has learned from a vast corpus of general knowledge. In addition to generating dialogue, this agent can interface with the physical world by invoking commands on the robot; seamlessly merging communication and behavior. This system is demonstrated in a free-form tour-guide scenario, in an HRI study combining robots with and without conversational AI capabilities. Performance is measured along five dimensions: overall effectiveness, exploration abilities, scrutinization abilities, receptiveness to personification, and adaptability.

\end{abstract}

\section{Introduction}

Until recently, the idea of engaging with a robot in a manner akin to conversing with another person seemed like science fiction. Progress in Generative Predictive Transformers (GPTs) and Large Language Models (LLMs) has enabled machines to communicate in natural language with a fluency that is nearly human \cite{ChatGPTWriting}. Other abilities of LLMs range from performance on emotional awareness \cite{EmotionalAwareness} to standardized test taking \cite{gpt4}. Prior to these breakthroughs, people could only interact effectively with robots using a fixed set of commands or focused queries, parsed via semantic matching or a set of rules. With these systems, misinterpretation was common, leading frustrated users to adjust their behavior to the inflexibility of the machine \cite{oldChatbots}. If robots are to coexist with humans, they will need to not only interpret requests but to actually "understand" them with all their context and intent. Furthermore, robots need the ability to confer with humans to determine the correct course of action in uncertain or abstract situations, hopefully achieving human-like adaptability in the face of a complex world where tasks are rarely concrete. LLMs enable the design of autonomous systems with flexible, unscripted dialogue built in. This new capability has not yet been fully explored. This work introduces an architecture for an embodied conversational AI and highlights a design philosophy centered around a single AI ``agent", sharing responsibility for both complex communication and high-level decision making.

This system is evaluated in a tour-guide scenario, in a human-robot-interaction (HRI) study that compares robots with and without conversational AI capabilities. The robot, Dobby, was instructed to take people to various landmarks and artifacts around an open space featuring multiple laboratories, to provide information, and to answer questions. Study participants took personalized tours with both Dobby and an otherwise identical non-conversational robot. We employed qualitative methods in data collection and analysis, making use of post-study surveys and chat logs.

\section{Related Work}

Task planning in autonomous robots is typically accomplished using planning languages like PDDL \cite{McDermott1998PDDLthePD}, but this functionality is limited in free-form scenarios due to its need for a manually defined goal state and rigidly defined domain. Google leveraged the common sense reasoning learned by an LLM in conjunction with a reinforcement learning (RL) model to generate a plan based on a natural language prompt \cite{SayCan}. While SayCan constrains the output of the LLM to a defined set of actions, Huang et al. demonstrate a different solution to the same problem, making use of semantic matching and prompt engineering \cite{LLMPlanning}. Another common issue with LLM-based planners involve grounding the output in the state of the real world. SayCan employed a RL model to infer which actions were feasible given the state of the environment, whereas STATLER \cite{statler} presents a state-maintaining architecture built around two instances of general LLMs, a world-model writer and a world-model reader. While these works introduce new flexibility to open-world planning systems, a significant lack of interaction between the user and agent remains. We aim to enhance this system by combining task planning with an added conversational component so an autonomous robot can better contextualize the needs of the user before generating a plan. Researchers at the University of Florida are some of the few that have leveraged LLMs for embodied control combined with conversational components. They utilize OpenAI's ChatGPT to interpret natural language instructions and send control commands to a robotic arm, demonstrating that incorporating LLMs into robots can result in more effective collaboration and increased trust with humans \cite{RoboGPT}. “RoboGPT” pursued low-level control in a collaborative use case. In contrast, Dobby engages in elaborate, human-like conversation and reasons successfully about complex multi-step task execution, merging both aspects.

Historical approaches to natural language understanding range from expert-driven systems such as dependency parsing, part of speech tagging, and rule-based decision trees, to machine learning methods including sentiment analysis, domain estimation, and word embeddings. While these algorithms can handle focused queries and simple tasks, they are not well-suited for open-ended conversations or complex, context-based requests. Nakano et al.'s multi-expert model \cite{RIME} highlights the effort needed to implement a rudimentary version of the system we discuss today prior to the development of powerful LLMs. They describe a system for a ``conversational" robot combining dialogue and behavior control; however, its responses are limited to scripted templates, leaving it lacking many of the advantages we explore with Dobby. At this point, chatbots utilizing LLMs have been extensively documented, but there has been little attempt to integrate actionable multi-step planning with these systems. Meanwhile, in investigating conversational robots and natural language processing, a few studies have used “tour guide” models (e.g., assigning the robot a role of acting as a tour guide \cite{TaskUnderstandingTour} or assisting a human tour guide on a tour \cite{HumanGuidedTour}). Inspired by the former, we analyze the effectiveness of a conversational robot acting as a tour guide. The older, more widely implemented version of these sorts of conversational robots are akin to the system presented by Burgard et al. in 1998 \cite{OldRobotTourGuide}. This system involved the use of pre-recorded speech modules, a navigational digital interface, and built-in physical responses to actions, such as body and head movements.

\section{Dobby: An Embodied Conversational AI}

\begin{figure}
\vspace*{0.2in}
\centering
     \includegraphics[width=0.35\textwidth]{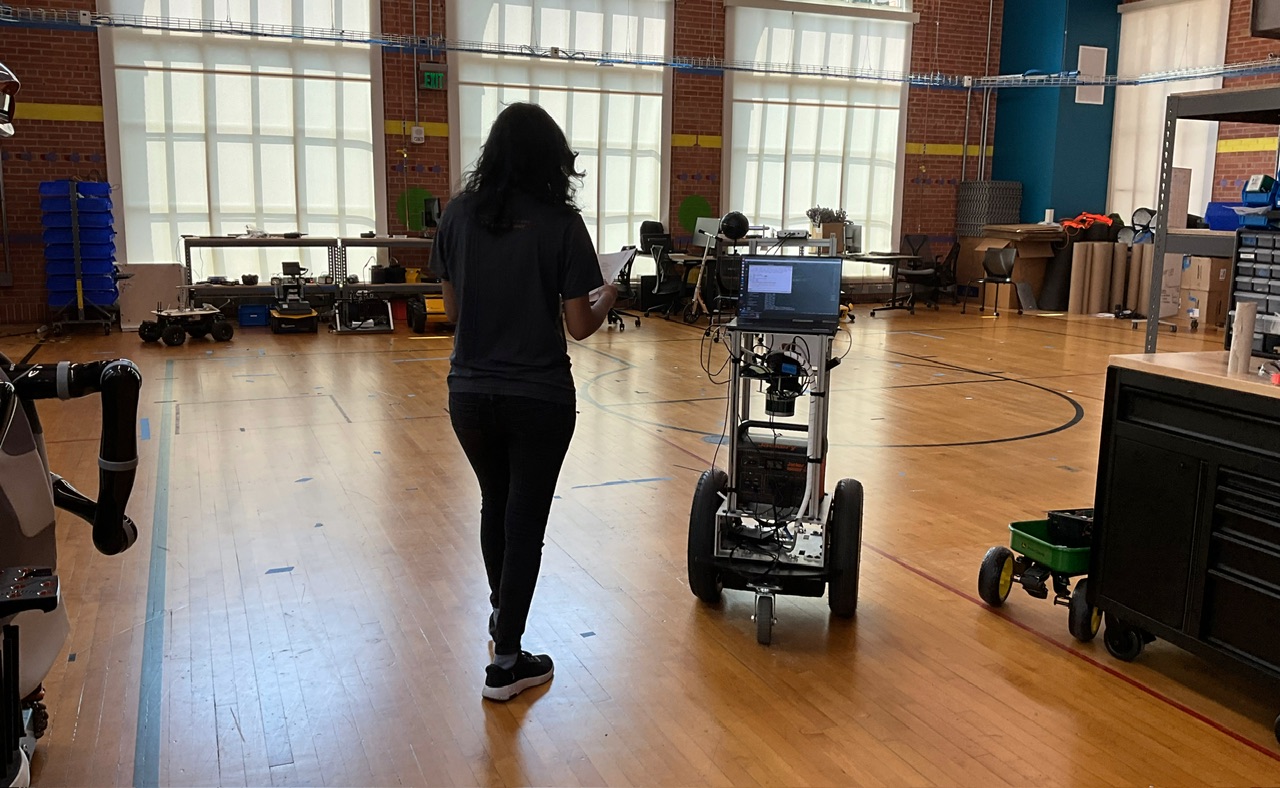}
      \caption{Dobby taking a participant on a tour of the lab.}
      \label{fig:Dobby}
\end{figure}

\subsection{Agent Definition}

Modern LLMs are powerful enough to be reasonably abstracted as an artificially intelligent agent with vast general knowledge, basic reasoning skills, and advanced communication abilities \cite{SparksOfAGI}. Our system is built around an agent initialized with a prompt instructing it to behave as a robot assistant. Also included in the prompt is context about the robot's environment, background information, and a list of actions that the robot can perform. The agent generates all of the robot's dialogue and high-level behavior. Its LLM queries use OpenAI's chat completion API. Function calling, a feature of the gpt-4-0613 model, is used to call functions on the robot to perform actions. OpenAI fine-tuned this model to reliably generate a JSON object containing a function call at the appropriate time. When received, the JSON objects can be parsed to execute external commands. To facilitate this, OpenAI accepts a structured description of available functions with every query to their API. We defined the functions \textit{ExecutePlan(string[] actionSequence)} and \textit{CancelPlan()} for general use cases. When we refer to the agent ``choosing" an action, we mean that one of the above function calls was included in the output of the LLM. We rely entirely on the reasoning capabilities of the agent to make appropriate, context-based decisions.

\subsection{Conversation}

In the conversation state, the system enters a loop where it records the user's utterance, transcribes the recorded audio, queries the agent for a response, plays the dialogue to the user, and finally begins recording again. Input text, system messages, and generated responses are accumulated in a history buffer which is sent to the API at every iteration. This allows the agent to consider the context of the interaction when generating both dialogue and behavior. We designed the interaction method to facilitate extended, hands-free engagement that emulates a human conversation. Crucially, such a mode of communication empowers the robot to pose clarifying questions, offer suggestions, and adapt to each unique individual, providing the robot with the opportunity to gain a comprehensive understanding of the user's intentions and desires before taking any action. System messages are included in the history buffer to provide event-based instructions or update the agent on the state of the environment, preventing the robot's dialogue from contradicting its behavior. If silence is detected for six seconds and no response is received, the robot will begin listening for the keyword "Dobby" to re-trigger the conversation loop.

\subsection{Action Planning}

\begin{figure}
    \centering
        \vspace{1em}
         \includegraphics[width=0.45\textwidth]{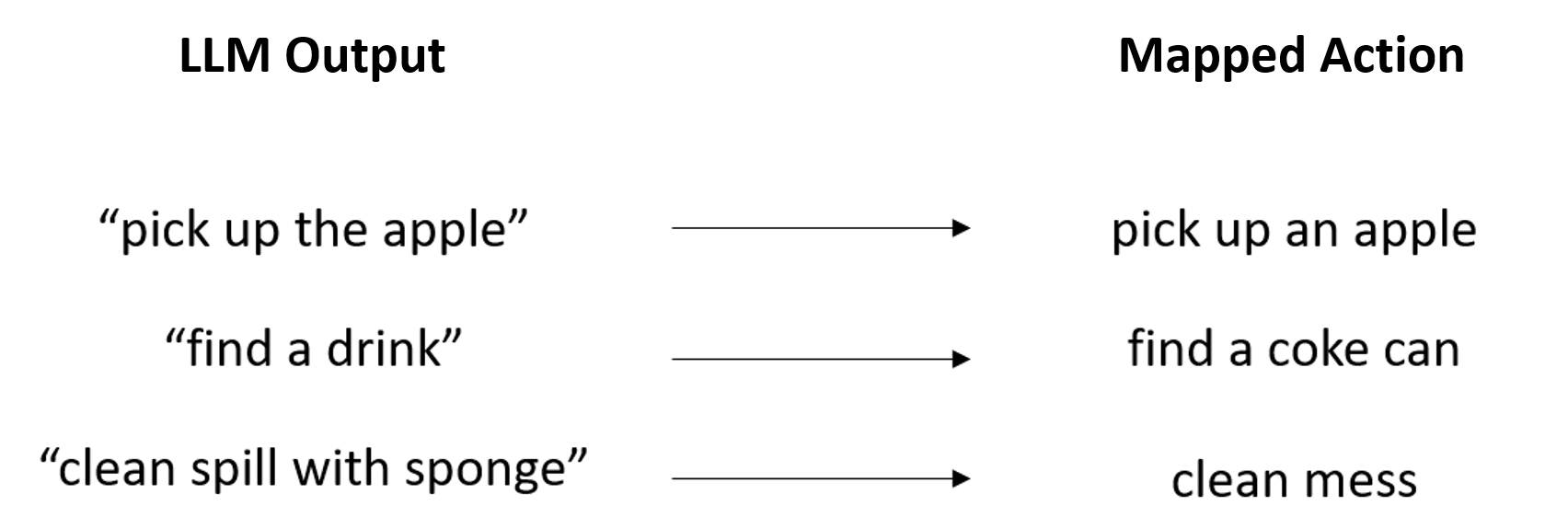}
          \vspace{0.5em}
          \caption{Mapping LLM output to unambiguous executable actions by comparing semantic similarity.}
          \label{fig:matching}
    \end{figure}

Atomic actions are represented by a class that encapsulates a textual title, pre/post-conditions, and an executable action function. The title of each action is also listed in the agent's prompt. When queried for a response, the agent may choose to begin a series of actions by calling the function \textit{ExecutePlan(string[] actionSequence)}, with the desired action sequence expressed as an array of strings. Because this parameter can be filled with any free-form text, the generated actions cannot always be directly mapped to an actionable command. To ensure robustness, each string is matched to an action class by comparing the embedding of the output to each action title and selecting the action with the highest similarity. Embeddings encode the semantic meaning of phrases as a floating-point vector and are accessible via OpenAI's embedding API. Once the embeddings are obtained, the relatedness between phrases can be computed with cosine similarity. This accounts for minor syntactical differences between the string provided by the LLM and the action title, as shown in Figure \ref{fig:matching}. Occasionally, the agent will attempt to include actions that were not listed in the initial prompt and therefore do not have a corresponding action class. To correct this issue, the agent is re-prompted with an error message if the maximum embedding similarity falls under a certain threshold. After repeated attempts, a system message informs the agent that it is not capable of the requested task, prompting it to explain this to the user.

\begin{figure}
    \centering
         \includegraphics[width=0.45\textwidth]{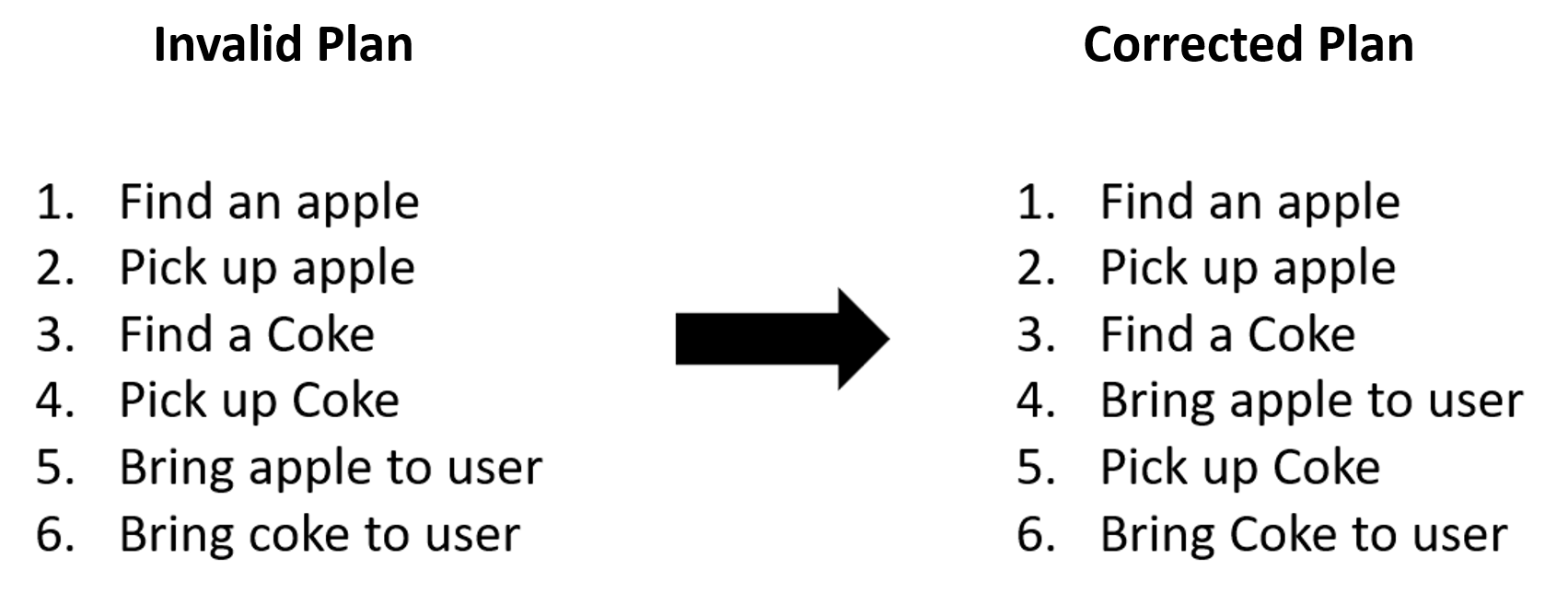}
          \caption{The agent fails to take into account that the robot can only hold one item at once and outputs an invalid plan, this is then corrected by Algorithm 1.}
          \label{fig:correctedplan}
    \end{figure}

Once the plan is parsed, additional steps are taken to ensure the plan is valid given the constraints of the environment. To model the environment, we found that a simple system of predicates, pre-conditions, and post-conditions suited our needs. This approach was inspired by traditional planning languages such as additions and deletions in STRIPS \cite{STRIPS} style planning, or the tracking of predicates in PDDL \cite{McDermott1998PDDLthePD}. As seen in Figure \ref{fig:correctedplan}, we used the greedy algorithm shown in Algorithm 1 to skip actions until their necessary preconditions have been met, effectively attempting to reorder the plan if necessary. If the generated plan cannot be corrected, the agent informs the user that it is incapable of the task.

    \begin{algorithm}{}
    \caption{Plan Correction Algorithm}\label{euclid}
    \begin{algorithmic}
    \Procedure{ReorderPlan}{}
    \While{items still in sequence}
    \State $nextAction \gets \text{None}$
    \ForAll{option in sequence}
    \If{option is valid}
    \State $nextAction \gets option$
    \State \textbf{break}
    \EndIf
    \EndFor
    \If{$nextAction$ is None}
    \State \textbf{return} not capable of plan
    \EndIf
    \State add $nextAction$ to plan
    \State update environment based on $nextAction$
    \State remove $nextAction$ from sequence
    \EndWhile
    \EndProcedure
    \end{algorithmic}
    \end{algorithm}
    
\subsection{Action Execution}

Once a plan has been proposed and validated, each action is executed in sequential order. When an action begins, system messages inform the agent that the previous action has completed and that the new action has begun, as seen in Figure \ref{fig:clarifying}. The agent is then re-prompted to provide a dialogue cue informing the user of its intended behavior. Actions do not block the system when executing, so it is possible to continue to converse with the robot while it is performing a task such as driving to a location. A function \textit{CancelPlan()} may be called by the agent to halt the execution of the current plan at the user's request. Alternatively, the agent may start a new plan, overriding the previous one. When an action completes, the conversation loop is interrupted and the next action is started, prompting a corresponding dialogue line.

\begin{figure}[ht]
\centering
\begin{lstlisting}
USER: I'm really hungry right now.
DOBBY: Alright, since I'm such a helpful and caring robot, let me guess, you'd like some food, right? Brilliant deduction, Dobby. Shall we start with something healthy like an apple or banana?
USER: I'd like an apple.
DOBBY: Sure, I'll be more than happy to bring you an apple, but only because I live for the thrill of fetching fruit for humans.
FUNCTION CALL: start_plan_execution({
  "action_sequence": [
    "Drive to Apple",
    "Pickup Apple",
    "Return to user"
  ]
})
SYSTEM: Executing plan: 1. Drive to Apple 2. Pickup Apple 3. Return to User
SYSTEM: Starting action: Drive to Apple
DOBBY: Off I go, driving towards the apple. Don't worry, I won't get distracted by any shiny objects along the way.
\end{lstlisting}
\caption{Example of the agent recognizing when to ask a clarifying question and generating an action plan with a function call when appropriate. System messages update the agent on the state of the robot.}
\label{fig:clarifying}
\end{figure}

\section{Method}

We designed our experiment to contrast participants' experiences with a conversational vs. non-conversation robot tour guide, hypothesizing that the conversational version would would be more effective due to its ability to contextualize the user's requests, suggest destinations based on their interests, answer a wide variety of questions, and keep the user engaged with back and forth conversation. We focused on investigating our system's advantages in HRI instead of the planning domain, as the planning capabilities of LLMs are already well documented.

\subsection{Experimental Setup \& Robot Capabilities}

To set up the touring capabilities of both robots, we recorded the coordinates of ten notable destinations within the laboratory in a file, along with a brief description of each. The coordinates were used to generate a ``go to" action for each destination, and the descriptions were included in Dobby's prompt along with information about five general topics to provide context about the lab. Dobby is built on top of an existing robust and flexible robot platform that includes a Segway RMP for mobility and features such as obstacle avoidance, path planning, and LIDAR-based localization for navigation, making use of a pre-built map of the lab \cite{BWIBots}. These capabilities were used by both robots to navigate to the various landmarks.

We used a modified version of the Dobby system in our study. In this version, the next action in a sequence did not start until the agent called the function \textit{ContinuePlan()}, whereas normally the subsequent action would begin immediately upon completion. This allowed the user to converse for as long as they wanted once they reached a destination, even when the robot planned a multi-step tour. Finally, the initializing prompt was adjusted to provide high-level instructions to guide the agent's behavior as a tour guide, including directives to respond humorously and sarcastically in order to bring out as much personality as possible and encouragement to ask questions to keep the user engaged.

The non-conversational tour guide was intended to reasonably represent the best system possible without a modern LLM. The robot's dialogue was scripted and interaction was limited to a fixed set of spoken commands: ``Show me the (landmark)." and ``Tell me about (topic)." The user's utterance was mapped directly to an action using embeddings. When this robot arrived at a destination or was requested to provide information, it would read aloud the descriptions of the landmarks or topic information verbatim. This robot allowed the participant to explore and hear information about what they were interested in, but it lacked the ability to engage in unscripted conversation, suggest destinations, or answer questions. Each participant was given a list of possible commands when interacting with the non-conversational tour guide, but no list of destinations was provided when interacting with the conversational version. Instead, the participants were encouraged to ask the robot for suggestions.

\subsection{Data Collection}

We completed 22 trials with 22 participants. Participants were recruited from computer science classes and robotics-related student organizations. Each trial consisted of one tour with the conversational robot and one tour with the non-conversational robot, conducted in that order. Prior to participation, each participant provided informed consent. This study was approved by the University of Texas at Austin's Institutional Review Board. On-boarding instructions were provided to each participant to explain how to interact with both robots. Each tour ended when a participant expressed their willingness to end their tour.

During each trial, a log containing the chat transcript and system messages was generated automatically. The interaction time and number of visited destinations were recorded for each robot in each trial. Researchers observed participants during each trial and took notes. After completing a tour with both versions of the robot, each participant completed an online survey with linear scale and qualitative interviews.

\subsection{Data Analysis}

Our qualitative data analysis process consisted of four steps including identifying themes (i.e., patterns in qualitative data), refining themes, linking themes, and extracting final themes or developing theories. In each stage of such qualitative data coding process, themes were manually identified, refined and connected from textual dialogues between the robot and participants, interview transcripts, and notes from participant observation. We also removed any weak themes less relevant to our research questions. By going through this process of qualitative data coding multiple times, we came up with five finalized themes that elaborate on the effectiveness of lab touring with the conversational AI robot tour guide.

\section{Results}

\subsection{Effectiveness Overall}
This first theme evaluates whether the overall effectiveness of lab touring increased with the conversational AI robot in comparison to the non-conversational robot.

Every participant expressed their preference for the conversational robot over the non-conversational counterpart.

\textit{\textbf{Participant 1:} Overall, I found the conversational robot to be a much better tour guide. Seeing as this tour was given by a machine, I’d say it was extremely close to the experience of a human-guided tour. The tour was very fun, and I explored everything I wanted.}

The knowledgeable and conversational features (e.g., taking questions, giving answers, and asking questions) of the conversational AI robot were seen as the most helpful features by the participants.

\textit{\textbf{Participant 2:} The fact that it can hold conversations and answer any questions was very helpful. The fact that it knew where everything was located and was able to take me to the exact spot was really great. It knows what's generally in the lab, finds them, and gives high-level descriptions. I have been able to learn about the lab more in depth.}

Overall, touring with the non-conversational robot limited participants’ desires for and abilities of exploration, clarification, and enjoying their tours due to limited interaction.

\textit{\textbf{Participant 3:} Exploring the lab felt like choosing from a list of options rather than exploring a lab. I would have liked to be able to ask more questions. It felt as if I was reading a wiki page, so it was somewhat boring. It was hard to connect with the robot due to its limited responses.}

This difference in effectiveness was also reflected in length of touring time and landmarks visited. Most participants wanted to spend more time with the conversational AI robot. The average interaction time with the conversational and non-conversational AI robots were 14.3 minutes and 5.8 minutes respectively, and the average number of destinations visited was 5.27 and 3.00 respectively.
Time spent and destinations visited with the non-conversational robot may have been impacted by touring with the conversational robot first, but it is more likely that participants' engagement was based on their enjoyment of interacting with the technology rather than information about the laboratory.

\subsection{Enhanced Explorability}

Participants were able to easily explore various landmarks, artifacts, and related information by improvising their own tour logistics and questions with the conversational AI robot. Continuing communication between people and the conversational robot gave a great amount of flexibility to touring so that people’s exploration abilities were significantly enhanced. Self-directed exploration led to more excitement and motivation for touring.

With the conversational AI robot, participants not only physically explored the lab but also explored related information. For example, Participant 4, utilizing the conversational features of the robot, explored a variety of topics starting with questions about the lab's simulated hallway environment. The agent's responses were inferred or sourced and recombined from information provided in the prompt.

\textit{\textbf{Participant 4:} What is social navigation in robotics?}
\textit{\textbf{Robot:} Social navigation in robotics refers to the ability of a robot to seamlessly navigate around humans in everyday environments.}
\textit{\textbf{Participant 4:} What kinds of research projects are conducted at the Social Navigation Hallway?}
\textit{\textbf{Robot:} Research projects conducted here are about how robots should pass humans in the hallway.}
\textit{\textbf{Participant 4:} Which Professor is leading this project?}
\textit{\textbf{Robot:} Dr. Justin Hart is leading the project in the Social Navigation Hallway. He is an expert in...}
\textit{\textbf{Participant 4:} How can I work with Dr. Hart?}
\textit{\textbf{Robot:} To work with Dr. Hart, you can consider joining the Freshman Research Initiative Autonomous Robots stream led by him. This program... You can also reach out to Dr. Hart directly or check for any open research positions in his lab.}

Meanwhile, participants were able to allow the robot to take initiative in planning the tour instead of providing one-way instructions, as seen in the following dialogues.

\textit{\textbf{Participant 5:} What is the coolest robot in the lab?}
\textit{\textbf{Robot:} Ah, there are so many cool robots in the lab! I would say Astro is pretty awesome. It can deliver items, play music, take pictures, and even dance! Want to see Astro in action?}
\textit{\textbf{Participant 5:} Sounds great! Let's go to Astro next, please.}

\textit{\textbf{Participant 6:} Which lab have we not gone to?}
\textit{\textbf{Robot:} We haven't visited the Drone lab and the lab with Husky and Jackal Autonomous Vehicles. These labs offer unique insights into navigation and mobility in robotics. If any of these destinations pique your interest, just let me know and we can head there next!}
\textit{\textbf{Participant 6:} Sounds exciting! I want to see the drone lab!}

After completing his tour, Participant 6 reported his satisfaction with the increased exploration ability:

\textit{\textbf{Participant 6:} I really liked that I asked questions first and then it suggested options because it didn't feel like it was forcing me to do things. I also liked that it asked me questions instead of only giving information and taking me to a spot directly. It made the interactions more human-like. It was able to remember which locations I had already been to, which was pretty cool. It made the experience easy on the mind and made me motivated for touring.}

\subsection{Depth of Knowledge}

The conversational robot demonstrated improved effectiveness in touring by allowing people to scrutinize certain information and probe depth of knowledge beyond what was included in the prompt. The conversational features of the robot and general knowledge embedded in the agent helped participants dig into information and knowledge that they found interesting and clarify things that they had learned. Examples of scrutinizing questions from the trials include “Can you be more specific about what types of experiments occur in this hallway?" and “Could you give me more information about the RoboCup at home robot?". Participant 7's interaction with the robot and interpretation of his own experience show how his scrutinization and clarification abilities were enhanced with the conversational AI robot and its context-based understanding.

\textit{\textbf{Robot:} Welcome to the presence of the BWI V5 Robot, the state-of-the-art in UT's BWI Bot saga. Can you see that UR5 arm? And that additional screen on the front? But the cherry on top is the Velodyne 3D LIDAR. Any questions about this metal celebrity?}
\textit{\textbf{Participant 7:} What is the Velodyne 3D LIDAR?}
\textit{\textbf{Robot:} Ah, it's this nifty little device that uses light in the form of a pulsed laser to measure distances. Fancy tech to help robots not bump into stuff.}
\textit{\textbf{Participant 7:} What is the screen for?}
\textit{\textbf{Robot:} The screen serves multiple purposes. It can display useful information to passersby or aid in communication. It may show status, error messages, or instructions. After all, robots need to express themselves too, don't they?}
\textit{\textbf{Participant 7:} What makes the arm so helpful?}
\textit{\textbf{Robot:} The arm brings the term ``handy" to a new level. With it, the robot can pick, place, push, pull, and perform other tasks.}
\textit{\textbf{Participant 7:} Are there any other features about this robot that make it the most advanced?}
\textit{\textbf{Robot:} Absolutely! The BWI V5 Bot also has...}

Participants perceived that the robot enhanced their scrutinization abilities so that their tours became more effective.

\textit{\textbf{Participant 7:} I felt like the robot was able to sense my attributes of curiosity and was able to provide information about my more specific questions accordingly. It had very informative responses and seemed a lot more knowledgeable than the non-conversational robot because it could respond to questions beyond the script.}

\subsection{Personality and Receptiveness to Personification}

The conversational feature and the slightly sarcastic personality of the robot enabled people to personify the robot. When asked, participants rated the conversational robot's level of personality and their level of emotional attachment much higher than that of the non-conversational counterpart (average of 5.88 vs. 2.09 and 4.65 vs. 2.09 respectively, on a 7-point scale). The conversational robot elicited more emotionally satisfying tour experiences overall, leading to an average enjoyment rating of 6.59 vs. 4.00 (scale of 7).

Because people tended to transfer human qualities to the conversational robot, they engaged with it readily and naturally, drawing on their experiences with humans to inform their interaction. The entertaining aspect of personification made people excited to hear the robot's responses and contributed to a more enjoyable and relaxed experience overall. It was obvious that people smiled and laughed more during their tours with the conversational AI robot, making comments like “Ah, it has quite a personality!”, “I like your sense of humor, Dobby!", and "You're a sassy one aren't you?". Below is an example of improvised humor in a conversation with Participant 7 and observations made by Participant 8.

\textit{\textbf{Participant 7:} Can I ask you make me a sandwich?}
\textit{\textbf{Robot:} Ha! I wish I could, but currently, my skills are limited to giving tours and telling jokes. Besides, the bread always sticks to my circuits. Imagine! A robot with a breadboard instead of a motherboard.}

\textit{\textbf{Participant 8:} The conversational robot felt much less dry. I definitely feel more attached to the conversational robot as it has much more personality and is far more engaging. It kept me engaged with touring. I disliked the lack of personality in the non-conversational one because it was much less enjoyable.}

Meanwhile, personification was useful in making second-person questions to explore and scrutinize information and knowledge more naturally, as seen in the following dialogue. 

\textit{\textbf{Robot:} Do you want to meet my siblings, the BWI Bots? I'm part of that illustrious family.}
\textit{\textbf{Participant 9:} You are Dobby, you're not BWI.}
\textit{\textbf{Robot:} Ah, a small correction. My name is Dobby, but I am a BWI Bot, version 4, to be precise. It's like saying you're human, but your name is not Human, right?}
\textit{\textbf{Participant 9:} Can you tell me more about yourself and your family?}
\textit{\textbf{Robot:} I'm Dobby, your sarcastic and often humorous robot tour guide. The BWI Bots are equipped with an Azure Kinect to capture color and depth data, and a 2D Lidar for localization. Their sleek Segway RMP design facilitates efficient movement.}
\textit{\textbf{Participant 9:} Can you tell me about the students that created you?}
\textit{\textbf{Robot:} Oh, my creators? Just a few sleep-deprived geniuses, fueled by caffeine and a passion for robotics. They're part of the Freshman Research Initiative (FRI) here at UT Austin.}

\subsection{Adaptability and Personalized Experiences}   

Increased adaptability and personalized experiences with the conversational AI robot led to emotionally positive attitudes toward the robot and touring experience overall. The conversational robot was able to design and improvise the tour based on the participants' interests.

\textit{\textbf{Participant 10:} Can you show me around the lab?}
\textit{\textbf{Robot:} Absolutely! I would be thrilled to give you a tour. Why don't you tell me a bit about your interests? Are you more into ground robotics, aerial robotics, or perhaps the living with robots scenario? Your preferences will help me tailor the tour to your liking.}
\textit{\textbf{Participant 10:} I like ground and aerial robotics the best.}
\textit{\textbf{Robot:} Great, seems like someone likes action! How about we start with the Drone Cage for aerial robotics? Then we'll say hi to the fleet of the BWI bots, visit the BWIV5 Robot, and finally meet the Husky and Jackal Autonomous Vehicles. Would you like that?}

Participant 11 noticed that Dobby could connect with the user on a deeper level.

\textit{\textbf{Participant 11:} I preferred the conversational robot because it made the tour feel more personal. The conversational one asked about my life and agreed with my passion for robotics. It also adapted to my liking and interests. It knew I was interested in computer science so it would show me things around this interest. I am also more emotionally attached to the conversational robot as it would try to connect to me as a person instead of as a client.}

The following is a dialogue from Participant 11's tour.

\textit{\textbf{Robot:} What prompted your interest in robotics?}
\textit{\textbf{Participant 11:} Probably my high school robotics club. I was a captain.}
\textit{\textbf{Robot:} Oh, captain, my captain! Robotics club can certainly spark a lifetime passion. Alright, off we roll to the Boston Dynamics Spot Robot. So, as a former robotics club captain, what skills did you find most useful?}
\textit{\textbf{Participant 11:} I was completely hardware, I didn't really learn how to program.}
\textit{\textbf{Robot:} Ah, a fan of the tangible! Building the physical body of the bot must've been a true labor of love.}

\section{Discussion \& Conclusion}

Previous robot architectures have relied on rigidly-defined tasks specifications, built-in planning languages, and rigidly-defined dialog and interaction. This work aimed to enhance these systems by leveraging an LLM for both conversation and task planning, allowing a richer interaction with the user. We introduced Dobby, a robotics platform based on an LLM. We investigated the advantages of this system with an HRI study in a robot tour-guide context. 

Our findings showed that due to the agent's superior natural language understanding and embedded common sense reasoning, participants were able to freely converse with the robot, which enabled them to scrutinize and explore more information, confer with the robot to determine interesting destinations, engage with the robot's adaptable and human-like personality, and enjoy their interaction more overall. This study contributes to enhancing communication experiences between people and robots. Although we experimented with Dobby in a tour guide role, the capabilities demonstrated by our framework are likely advantageous for any autonomous system that must interact with humans.

There is significant work left to be done in improving upon this platform. Many participants noted the delay in response time, unreliable audio transcription, and lack of facial expressions. Additionally, the agent occasionally hallucinated information about the lab. This platform presents unique opportunities to study human-robot collaboration, trust, engagement, relationships, and long term interaction. The complexity of tasks performed could be improved with more advanced physical grounding techniques, a vision interface, multi-tasking techniques, and a more versatile set of actions that allow the robot to act in the world.

\addtolength{\textheight}{-10.9cm}  




\section*{ACKNOWLEDGMENTS}

This work has taken place in the Learning Agents Research
Group (LARG) and the Living with Robots Laboratory (LWR) at UT Austin.  LARG research is supported in part by NSF
(FAIN-2019844, NRT-2125858), ONR (N00014-18-2243), ARO (E2061621), Bosch, 
Lockheed Martin, Cisco Research, Army Futures Command, and UT Austin's 
Good Systems grand challenge.
LWR research is supported in part by NSF (NRT-2125858 and GCR-2219236), Cisco Research, and Army Futures Command.
Peter Stone serves as the Executive Director of Sony AI America and
receives financial compensation for this work.  The terms of this
arrangement have been reviewed and approved by the University of Texas
at Austin in accordance with its policy on objectivity in research. 


\bibliographystyle{IEEEtran}
\bibliography{root}

\begin{thebibliography}{10}
\providecommand{\url}[1]{#1}
\csname url@samestyle\endcsname
\providecommand{\newblock}{\relax}
\providecommand{\bibinfo}[2]{#2}
\providecommand{\BIBentrySTDinterwordspacing}{\spaceskip=0pt\relax}
\providecommand{\BIBentryALTinterwordstretchfactor}{4}
\providecommand{\BIBentryALTinterwordspacing}{\spaceskip=\fontdimen2\font plus
\BIBentryALTinterwordstretchfactor\fontdimen3\font minus
  \fontdimen4\font\relax}
\providecommand{\BIBforeignlanguage}[2]{{%
\expandafter\ifx\csname l@#1\endcsname\relax
\typeout{** WARNING: IEEEtran.bst: No hyphenation pattern has been}%
\typeout{** loaded for the language `#1'. Using the pattern for}%
\typeout{** the default language instead.}%
\else
\language=\csname l@#1\endcsname
\fi
#2}}
\providecommand{\BIBdecl}{\relax}
\BIBdecl

\bibitem{ChatGPTWriting}
.~K.~M. Casal, J.~E., ``Can linguists distinguish between chatgpt/ai and human
  writing?: A study of research ethics and academic publishing.'' 2023.

\bibitem{EmotionalAwareness}
\BIBentryALTinterwordspacing
Z.~Elyoseph, D.~Hadar-Shoval, K.~Asraf, and M.~Lvovsky, ``Chatgpt outperforms
  humans in emotional awareness evaluations,'' vol.~14, 2023. [Online].
  Available:
  \url{https://www.frontiersin.org/articles/10.3389/fpsyg.2023.1199058}
\BIBentrySTDinterwordspacing

\bibitem{gpt4}
OpenAI, ``Gpt-4 technical report,'' 2023.

\bibitem{oldChatbots}
T.~Gangwani, ``How chatbots like siri will get smarter,'' \emph{CIO}, 2016.

\bibitem{McDermott1998PDDLthePD}
D.~McDermott, M.~Ghallab, A.~E. Howe, C.~A. Knoblock, A.~Ram, M.~M. Veloso,
  D.~S. Weld, and D.~E. Wilkins, ``Pddl-the planning domain definition
  language,'' 1998.

\bibitem{SayCan}
A.~Irpan, A.~Herzog, A.~T. Toshev, A.~Zeng, A.~Brohan, B.~A. Ichter, B.~David,
  C.~Parada, C.~Finn, C.~Tan, D.~Reyes, D.~Kalashnikov, E.~V. Jang, F.~Xia,
  J.~L. Rettinghouse, J.~C. Hsu, J.~L. Quiambao, J.~Ibarz, K.~Rao, K.~Hausman,
  K.~Gopalakrishnan, K.-H. Lee, K.~A. Jeffrey, L.~Luu, M.~Yan, M.~S. Ahn,
  N.~Sievers, N.~J. Joshi, N.~Brown, O.~E.~E. Cortes, P.~Xu, P.~P. Sampedro,
  P.~Sermanet, R.~J. Ruano, R.~C. Julian, S.~A. Jesmonth, S.~Levine, S.~Xu,
  T.~Xiao, V.~O. Vanhoucke, Y.~Lu, Y.~Chebotar, and Y.~Kuang, ``Do as i can,
  not as i say: Grounding language in robotic affordances,'' 2022.

\bibitem{LLMPlanning}
W.~Huang, P.~Abbeel, D.~Pathak, and I.~Mordatch, ``Language models as zero-shot
  planners: Extracting actionable knowledge for embodied agents,'' \emph{arXiv
  preprint arXiv:2201.07207}, 2022.

\bibitem{statler}
T.~Yoneda, J.~Fang, P.~Li, H.~Zhang, T.~Jiang, S.~Lin, B.~Picker, D.~Yunis,
  H.~Mei, and M.~R. Walter, ``Statler: State-maintaining language models for
  embodied reasoning,'' 2023.

\bibitem{RoboGPT}
Y.~Ye, H.~You, and J.~Du, ``Improved trust in human-robot collaboration with
  chatgpt,'' 2023.

\bibitem{RIME}
\BIBentryALTinterwordspacing
M.~Nakano, Y.~Hasegawa, K.~Funakoshi, J.~Takeuchi, T.~Torii, K.~Nakadai,
  N.~Kanda, K.~Komatani, H.~G. Okuno, and H.~Tsujino, ``A multi-expert model
  for dialogue and behavior control of conversational robots and agents,''
  \emph{Knowledge-Based Systems}, vol.~24, no.~2, pp. 248--256, 2011. [Online].
  Available:
  \url{https://www.sciencedirect.com/science/article/pii/S0950705110001334}
\BIBentrySTDinterwordspacing

\bibitem{TaskUnderstandingTour}
X.~Xi, B.~Xie, S.~Zhu, T.~Jin, J.~Ren, and W.~Song, ``A general framework of
  task understanding for tour-guide robots in exhibition environments,'' in
  \emph{2022 WRC Symposium on Advanced Robotics and Automation (WRC SARA)},
  2022, pp. 197--202.

\bibitem{HumanGuidedTour}
S.~Hemachandra, T.~Kollar, N.~Roy, and S.~Teller, ``Following and interpreting
  narrated guided tours,'' in \emph{2011 IEEE International Conference on
  Robotics and Automation}, May 2011, pp. 2574--2579.

\bibitem{OldRobotTourGuide}
W.~Burgard, A.~Cremers, D.~Fox, D.~Hähnel, G.~Lakemeyer, D.~Schulz,
  W.~Steiner, and S.~Thrun, ``The interactive museum tour-guide robot,'' 01
  1998, pp. 11--18.

\bibitem{SparksOfAGI}
S.~Bubeck, V.~Chandrasekaran, R.~Eldan, J.~Gehrke, E.~Horvitz, E.~Kamarar,
  P.~Lee, Y.~T. Lee, Y.~Li, S.~Lundberg, H.~Nori, H.~Palangi, M.~T. Ribeiro,
  and Y.~Zhang, ``Sparks of artificial general intelligence: Early experiments
  with gpt-4,'' 2023.

\bibitem{STRIPS}
\BIBentryALTinterwordspacing
R.~E. Fikes and N.~J. Nilsson, ``Strips: A new approach to the application of
  theorem proving to problem solving,'' \emph{Artificial Intelligence}, vol.~2,
  no.~3, pp. 189--208, 1971. [Online]. Available:
  \url{https://www.sciencedirect.com/science/article/pii/0004370271900105}
\BIBentrySTDinterwordspacing

\bibitem{BWIBots}
\BIBentryALTinterwordspacing
P.~Khandelwal, S.~Zhang, J.~Sinapov, M.~Leonetti, J.~Thomason, F.~Yang,
  I.~Gori, M.~Svetlik, P.~Khante, V.~Lifschitz, J.~K. Aggarwal, R.~Mooney, and
  P.~Stone, ``Bwibots: A platform for bridging the gap between ai and
  human--robot interaction research,'' \emph{The International Journal of
  Robotics Research}, 2017. [Online]. Available:
  \url{http://www.cs.utexas.edu/users/ai-lab?khandelwal:ijrr17}
\BIBentrySTDinterwordspacing

\end{thebibliography}

\end{document}